# Towards Alzheimer's Disease Classification through Transfer Learning

1st Marcia Hon
*Data Science*
Ryerson University
Toronto, ON
marcia.hon.29@ryerson.ca

2nd Naimul Mefraz Khan
*Department of Electrical & Computer Engineering*
Ryerson University
Toronto, ON
n77khan@ryerson.ca

*Abstract*—Detection of Alzheimer's Disease (AD) from neuroimaging data such as MRI through machine learning have been a subject of intense research in recent years. Recent success of deep learning in computer vision have progressed such research further. However, common limitations with such algorithms are reliance on a large number of training images, and requirement of careful optimization of the architecture of deep networks. In this paper, we attempt solving these issues with transfer learning, where state-of-the-art architectures such as VGG and Inception are initialized with pre-trained weights from large benchmark datasets consisting of natural images, and the fully-connected layer is re-trained with only a small number of MRI images. We employ image entropy to select the most informative slices for training. Through experimentation on the OASIS MRI dataset, we show that with training size almost 10 times smaller than the state-of-the-art, we reach comparable or even better performance than current deep-learning based methods.

## I. INTRODUCTION

Alzheimers Disease (AD) is a neurodegenerative disease causing dementia in elderly population. It is predicted that one out of every 85 people will be affected by AD by 2050 [1]. Early diagnosis of AD can be achieved through automated analysis of MRI images with machine learning. It has been shown recently that in some cases, machine learning algorithms can predict AD better than clinicians [2], making it an important field of research for computer-aided diagnosis.

While statistical machine learning methods such as Support Vector Machine (SVM) [3] have shown early success in automated detection of AD, recently deep learning methods such as Convolutional Neural Networks (CNN) and sparse autoencoders have outperformed statistical methods.

However, the existing deep learning methods train deep architectures from scratch, which has a few limitations [4], [5]: 1) properly training a deep learning network requires a huge amount of annotated training data, which can be a problem especially for the medical imaging field where physician-annotated data can be expensive, and protected from cross-institutional use due to ethical reasons; 2) training a deep network with large number of images require huge amount of computational resources; and 3) deep network training requires careful and tedious tuning of many parameters, sub-optimal tuning of which can result in overfitting/underfitting, and, in turns, result in poor performance.

An attractive alternative to training from scratch is fine-tuning a deep network (especially CNN) through transfer learning [6]. In popular computer vision domains such as object recognition, trained CNNs are carefully built using large-scale datasets such as ImageNet [7]. Through Transfer Learning , these trained networks can be used with smaller datasets with only fine-tuning the final fully-connected layers of CNN. Transfer learning has been proven to be robust even for cross-domain application, such as networks trained on natural images used with medical images [8].

In this paper, we investigate the adaptation of two popular CNN architectures into an AD diagnosis problem through transfer learning, namely VGG16 and Inception. These two architectures are previous winners of the ImageNet Large Scale Visual Recognition Challenge (ILSVRC), a yearly competition that evaluates object detection in images [7]. With the two winners being open source, the architectures as well as the pre-trained weights are readily available. Although the architectures are trained on a different domain (natural images of ImageNet as opposed to MRI images in our paper), we show that the pre-trained weights generalize very well for AD diagnosis when the training data is picked intelligently.

Since our target was to test the robustness of transfer learning on a small training set, merely choosing training data at random may not provide us with a dataset representing enough structural variations in MRI. Instead, we picked the training data that would provide the most amount of information through image entropy. We show that through intelligent training selection and transfer learning, we can achieve comparable, or even better performance than training a deep network from scratch with minimal parameter optimization.

## II. RELATED WORKS

Classical machine learning methods such as SVM and feed-forward neural networks have been applied successfully to diagnose AD from structural MRI images [3], [9]. One such recent method is [9], where a dual-tree complex wavelet transform is used to extract features, and a feed-forward neural network is used to classify images. Elaborate discussion and comparative results with other popular classical methods can also be found in [9].

Recently, deep learning methods have outpeformed classical methods by a large margin. A combination of patches extracted from an autoencoder followed by convolutional layers for feature extraction were used in [10]. The method was further improved by using 3D convolution in [11]. Stacked autoencoders followed by a softmax layer for classification was used in [12]. Popular CNN architectures such as LeNet and the first Inception model were used in [13]. While these deep learning methods provide good accuracy results, as we show in the experiments, due to our intelligent training data selection and use of transfer learning, we get comparable, or even better accuracy while using only a fraction of the training data.

Transfer learning or fine-tuning has also been explored in medical imaging. [14] provides an in-depth discussion and comparative results of training from scratch vs fine-tuning on some medical applications. They show that in most cases, fine-tuning outperforms training from scratch. Fine-tuned CNNs have been used to localize planes in ultrasound images [15], classify interstitial lung diseases [8], and retrieve missing or noisy plane views in cardiac imaging [16]. All these methods prove that employing transfer learning in the medical imaging domain has tremendous value, and has the potential to achieve high accuracy in AD detection with smaller training dataset when compared to training from scratch.

## III. METHODOLOGY

### A. Convolutional Neural Networks and Transfer Learning

The core of Convolutional Neural Networks (CNN) are convolutional layers which can extract local features (e.g. edges) across an input image through convolution. Each node in a convolutional layer is connected to a small subset of spatially connected neurons. To reduce computational complexity, a max pooling layer follows convolutional layers, which reduces the size of feature maps by selecting the maximum feature response in a local neighborhood. Pairs of convolutional and pooling layers are followed by a number of fully-connected layers, where a neuron in one layer has connections to all activations in the previous layer. Fully-connected layers help learning non-linear relationship among the local features extracted by convolutional layers. Finally, a soft-max layer follows the fully-connected layers, which normalizes the outputs to desired levels.

CNNs are trained with the back-propagation algorithm, where in each iteration, weights associated with the neurons in the convolutional layers are updated in a way that minimizes a cost function. When training from scratch, the weights are typically initialized randomly, drawing from a normal distribution.

However, with a small training dataset, this may result in the cost function getting stuck in a local minima, which may result in overfitting/underfitting. A better alternative in such case is transfer learning, where pre-trained weights from the same architecture, but on a different larger dataset from the same/different domain is used to initialize the network layers. Only the last fully-connected layer is re-trained with the training data. This not only provides us with a robust set of pre-trained weights to work with, it gives us the opportunity to employ proven network architectures in our problem. The two popular architectures that we will use in this paper are:

1) **VGG16**:
   VGG16 is a 16-layer network built by Oxfords Visual Geometry Group (VGG) [17]. It participated in the ImageNet competition in ILSVRC-2014. One of the main reasons that VGG16 won the competition, is that it is one of the first architectures to explore network depth by pushing to 16-19 layers and using very small (3x3) convolution filters.

2) **Inception**:
   The Inception architecture is a variant of deep learning architecture built by Google. Throughout the years, there have been different iterations. Inception V4, discussed in this article, was introduced in [18]. The Inceptions' breakthrough is in the realization that non-linear functions can be learned by changing how convolutional layers are connected. Accordingly, the fully-connected layer is discarded in preference of a global average pooling that averages the feature maps and then connect with a softmax layer for classification. Thus, there are less parameters, and as a result, less overfitting.

### B. Most informative training data selection

While transfer learning provides an opportunity to use smaller set of training data, choosing the best possible data for training is still critical to the success of the overall method. Typically, from a 3D MRI scan, we have a large number of images that we can choose from. In most recent methods, the images to be used for training are extracted at random. Instead, in our porposed method, we extract the most informative slices to train the network. For this, we calculate the *image entropy* of each slice. In general, for a set of $M$ symbols with probabilities $p_1, p_2, \ldots, p_M$ the entropy can be calculated as follows [19]:

$$H = -\sum_{i=1}^{M} p_i \log p_i. \quad (1)$$

For an image (a single slice), the entropy can be similarly calculated from the histogram [19]. The entropy provides a measure of variation in a slice. Hence, if we sort the slices in terms of entropy in descending order, the slices with the highest entropy values can be considered as the most informative images, and using these images for training will provide robustness.

## IV. EXPERIMENTAL RESULTS

In this section, we provide experimental results on the two aforementioned models. We implemented our deep learning methods using Keras with a TensorFlow backend, while the training data selection method described in Section III-B was implemented in MATLAB. Architecture models for Tensor-Flow and pre-trained weights were downloaded from open source repositories of the models [1]. Our target was to differentiate AD patients from Healthy Control (HC) through analyzing MRI scan data through the CNN architectures.

*A. Dataset*

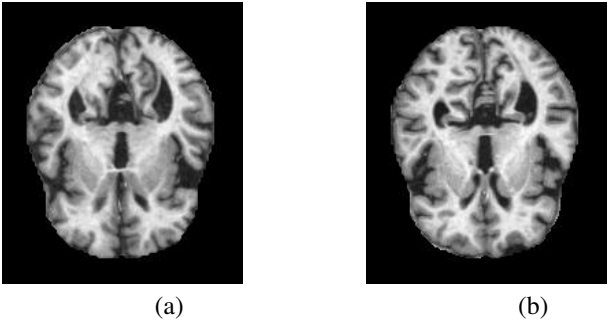

Fig. 1. Images from the OASIS dataset (a) AD. (b) HC.

We used structural MRI data from the Open Access Series of Imaging Studies (OASIS), accessible at (http://www.oasisbrains.org) [20]. Samples of MR brain images from the AD and the HC group are shown in Figure 1.

OASIS provides two types of data: cross-sectional and longitudinal. Since our target was to simply differentiate between AD and HC patients through the images, we used cross-sectional data. The dataset consists of 416 subjects whose ages are between 18 and 96. In our experiments, we randomly picked 200 subjects, 100 of whom were picked from the AD group, while the other 100 from the HC group. The grouping was determined by the Clinical Dementia Rating (CRD) variable ranging from 0 to 2. The assumption is that a 0 implies HC and those greater than 0 are AD.

To test the generalization power of transfer learning, we only used a limited number of images from these scans to train our networks, especially when compared to other CNN-based Alzheimer's detection algorithms (further comparison in Section IV-C). We used our entropy-based sorting mechanism to pick the most informative 32 images from the axial plane of each 3D scan. That resulted in a total of 6400 training images, 3200 of which were AD and the other 3200 were HC.

To be compatible with the pre-trained models of VGG16 and Inception V4, the images were resized to be 150X150 for VGG16, and 299X299 for Inception V4.

*B. Accuracy results*

5-fold cross-validation was used to obtain the results, with an 80% - 20% split between training and testing. To test the power of transfer learning, we also trained a VGG16 network from scratch for comparison. For the transfer learning models, pre-trained network weights from ImageNet were obtained. For VGG16, 100 epochs were used with a batch size of 40. For Inception V4, 100 epochs with a batch size of 8 was used.

For the VGG16 model (both from scratch and with transfer learning), the RMSProp optimization model was used, which uses an adaptive learning rate.

For the Inception V4 model, stochastic gradient descent optimization with a learning rate of 0.0001 was used [2].

These hyperparameter values were the recommended settings from the respective methods. There is a possibility that a hyperparameter search method such as grid search will result in further improvement, but as we see from the reported results below, the recommended settings already provide excellent results. Since grid search is computationally very expensive, we plan to explore this in future with cloud computing or additional hardware resources.

TABLE I
TESTED MODELS AND CORRESPONDING AVERAGE ACCURACY FROM 5-FOLD CROSS-VALIDATION. STANDARD DEVIATION OVER THE 5 FOLDS IN BRACKETS.

| Model | Avg. Acc. (st. dev.) (%) |
|---|---|
| VGG16 (from scratch) | 74.12 (1.55) |
| VGG16 (transfer learning) | 92.3 (2.42) |
| Inception V4 (transfer learning) | 96.25 (1.2) |

Table I shows the accuracy results (with standard deviation) for the three models. As can be seen, VGG16, when trained from scratch, results in poor performance. This can be attributed towards the small training size that we used, which can result in overfitting/underfitting. However, using pre-trained model for transfer learning, the accuracy significantly increases. Finally, using Inception V4 with transfer learning results in high accuracy. This signifies the value of transfer learning, where using even a relatively smaller training dataset can result in an excellent automated system for AD detection.

*C. Comparison with other methods*

In this section, we discuss the results in relation to other recent methods, both in terms of accuracy and training size. It is important to note that while the other methods may have used different datasets and/or different experimental setups, the results are still comparable, since all the methods discussed below report the results with structural MRI images, which have high level of similarity across datasets, especially when the images have been preprocessed and the brain have been registered and segmented in the published datasets already.

We compare our results with five recent methods, four of which are deep learning-based. For all these methods, the training size was calculated based on the reported sample size and the cross-validation/training-testing split (e.g. in our method, there are total 6,400 images; a 5-fold cross-validation/ 80% - 20% split therefore results in a training size of 5,120).

---

[1] https://github.com/flyyufelix/cnn_finetune

[2] Models, weights, dataset, and code available at https://github.com/marciahon29/Ryerson_MRP

TABLE II
COMPARISON WITH THE STATE-OF-THE-ART IN TERMS OF ACCURACY AND TRAINING SIZE.

| Model | Average Acc.(%) | Training Size (# of Images) |
|---|---|---|
| Wavelet + NN [9] | 90.06 | 3,629 |
| DeepAD (Inception) [13] | 98.84 | 46,751 |
| 3DConv [11] | 95.39 | 117,708 |
| Sparse autoencoder + conv [10] | 94.74 | 103,683 |
| Stacked autoencoders [12] | 87.76 | 21,726 |
| **Inception V4 (transfer learning)** | 96.25 | 5,120 |

As can be seen from Table II, our method outperforms every method except [13] in terms of accuracy. The most significant effect of transfer learning can be seen when comparing the training sizes. As we can see, only [9] used a smaller training set than ours, which is not a deep learning method, and therefore, does not require a large training set. While comparing with the deep learning methods, we can see that our training dataset is considerably smaller. Especially if we compare the training sizes between [13] and ours, both of which use the Inception architecture, we can see that our accuracy results are very close to theirs despite using a training size almost 10 times smaller (5,120 vs 46,751). This is due to the fact that we not only use transfer learning, but also employ an intelligent entropy-based method to pick the most informative training images. The combination of these two techniques results in very good performance with a comparatively smaller training dataset. A smaller training dataset is significant because it frees the method from dependence on large tediously annotated training data, and also, improves the training computational time significantly.

## V. CONCLUSION

In this paper, we propose a transfer learning-based method to detect AD from structural MRI images. We test two popular architectures, namely VGG16 and Inception V4. Through pre-trained weights from ImageNet and fine-tuning, we can only use a small number of training images to obtain highly accurate results. Moreover, we employ an intelligent entropy-based technique to select the training dataset in a way that represents the most amount of information within a small set. We test our models on images from the OASIS brain imaging dataset, where 6,400 images extracted from MRI scans of 200 subjects are used to train the models. As we show, our method provides performance comparable to the state-of-the-art despite having a training set many times smaller.

In future, we plan to test the models with other datasets, and employ grid search to optimize the hyperparameters of the proposed models further to achieve better results. We also plan to test the robustness of our models for three-way classification with AD, HC, and Mild Cognitive Impairment (MCI).

Keeping up with the spirit of reproducible research, all our models, dataset, and code can be accessed through the repository at: https://github.com/marciahon29/Ryerson_MRP .